\documentclass{article}

\usepackage[final]{neurips_2019}

\usepackage[utf8]{inputenc} 
\usepackage[T1]{fontenc}    
\usepackage{hyperref}       
\usepackage{url}            
\usepackage{booktabs}       
\usepackage{amsfonts}       
\usepackage{nicefrac}       
\usepackage{microtype}      
\usepackage{graphicx}
\usepackage{makecell}

\usepackage[]{natbib}
\usepackage{subcaption} 
\usepackage{amsmath}

\usepackage{algorithm}
\usepackage{algpseudocode}
\usepackage{tabularx}
\usepackage{amsmath,amsfonts,amsthm,amssymb,amsopn,bm}

\title{Traffic Sign Detection and Recognition for Autonomous Driving in Virtual Simulation Environment}

\author{
  Meixin Zhu\\
  University of Washington\\
  Department of Civil \& Environmental Eng.\\
  \texttt{meixin92@uw.edu} \\
  \And
  Jingyun Hu\\
  University of Washington\\
  Department of Civil \& Environmental Eng.\\
  \texttt{jingyun@uw.edu} \\
  \And
  Ziyuan Pu\\
  University of Washington\\
  Department of Civil \& Environmental Eng.\\
  \texttt{ziyuanpu@uw.edu} \\
  \And
  Zhiyong Cui\\
  University of Washington\\
  Department of Civil \& Environmental Eng.\\
  \texttt{zhiyongc@uw.edu} \\
  \And
  Liangwu Yan\\
  University of Washington\\
  Department of Mechanical Eng.\\
  \texttt{liangy00@uw.edu} \\
  \And
  Yinhai Wang\\
  University of Washington\\
  Department of Civil \& Environmental Eng.\\
  \texttt{yinhai@uw.edu} \\
}

\begin{document}
\maketitle


\textbf{Abstract}:
This study developed a traffic sign detection and recognition algorithm based on the RetinaNet. Two main aspects were revised to improve the detection of traffic signs: image cropping to address the issue of large image and small traffic signs; and using more anchors with various scales to detect traffic signs with different sizes and shapes. The proposed algorithm was trained and tested in a series of autonomous driving front-view images in a virtual simulation environment. Results show that the algorithm performed extremely well under good illumination and weather conditions. Its drawbacks are that it sometimes failed to detect object under bad weather conditions like snow and failed to distinguish speed limits signs with different limit values.

\section{Introduction}
Traffic sign recognition systems form an important component of Advanced Driver-Assistance Systems (ADAS) and are essential in many real-world applications, such as autonomous driving, traffic surveillance, driver safety and assistance, road network maintenance, and analysis of traffic scenes. In this study, we are required to identify the traffic signs that appear randomly in a virtual driving environment and return the corresponding recognition results in the order in which they appear.  The virtual simulation environment is accompanied by interference factors such as pedestrians and non-motor vehicles, and has diverse weather conditions (including illumination).



\section{Related Work}
\cite{arcos2018evaluation} did a excellent review of traffic sign detection methods. The concluded that the state-of-the-art methods include detection modules (Faster R-CNN \citep{ren2015faster}, R-FCN \citep{dai2016r}, SSD \citep{liu2016ssd}, and YOLO \citep{redmon2018yolov3}) combined with various feature extractors (Resnet V1 50 \citep{he2016deep}, Resnet V1 101, Inception V2 \citep{szegedy2017inception}, Inception Resnet V2, Mobilenet V1 \citep{howard2017mobilenets}, and Darknet-19). Below is a summary of several frequently used detection methods.

\begin{itemize}
    \item Faster R-CNN:
    Faster R-CNN has two networks: region proposal network (RPN) for generating region proposals and a network using these proposals to detect objects. The main different here with Fast R-CNN is that the later uses selective search to generate region proposals. The time cost of generating region proposals is much smaller in RPN than selective search, when RPN shares the most computation with the object detection network. Briefly, RPN ranks region boxes (called anchors) and proposes the ones most likely containing objects.
    \item YOLO:
    The input image is divided into an S x S grid of cells. For each object that is present on the image, one grid cell is said to be “responsible” for predicting it. That is the cell where the center of the object falls into.
    Each grid cell predicts B bounding boxes as well as C class probabilities. The bounding box prediction has 5 components: (x, y, w, h, confidence). Adding the class predictions to the output vector, we get a S x S x (B * 5 +C) tensor as output.
    \item SSD:
    The deep layers cover larger receptive fields and construct more abstract representation, while the shallow layers cover smaller receptive fields.
    In SSD, multiple boxes for every feature point are called priors, while in Faster R-CNN they are called anchors.
\end{itemize}

\section{Data Description and Evaluation Metric}
Data come from a competition called \href{https://www.datafountain.cn/competitions/339}{Traffic Sign Recognition for Autonomous Driving in Virtual Simulation Environment}. The competition provides a series of autonomous driving front-view images in a virtual simulation environment, where the traffic signs are marked.

\textbf{Training data:} RGB images with a total size of around 14 GB, and the corresponding labels with the format as shown in Table 1.

\begin{table}[H]
  \caption{Training images' label file format}
  \label{train}
  \centering
  \begin{tabular}{lll}
    \toprule
    Field     & Type  & Description  \\
    \midrule
    
    Filename	& String & Image filename\\
    x1, y1 & int & Upper left corner x and y coordinate\\
    x2, y2 & int & Upper right corner x and y coordinate\\
    x3, y3 & int & Bottom left corner x and y coordinate\\
    x4, y4 & int & Bottom right corner x and y coordinate\\
    Type & int & \makecell[cl]{1: parking lot; 2: yield to parking; 3: driving on the right;\\ 4: left or right turn; 5: Bus passage; 6: driving on the left; 7: slow down;\\ 8: driving through or right turn for motorized vehicles; 9: yield to pedestrians;\\ 10: roundabout; 11: driving through or right turn; 12: no bus access;\\ 13: motorcycles are prohibited; 14: Prohibition of motor vehicles;\\ 15: Prohibition of non-motor vehicles; 16: no honking;\\ 17: Driving through or turning on bypass; 18: 40 km/h speed limit;\\ 19: 30 km/h speed limit; 20: honking; 0: others} \\
    
    \bottomrule
  
  \end{tabular}
  \label{train}
\end{table}

A few examples of the training images are shown below.

\begin{figure}[H]
    \centering
    \includegraphics[width=1\textwidth]{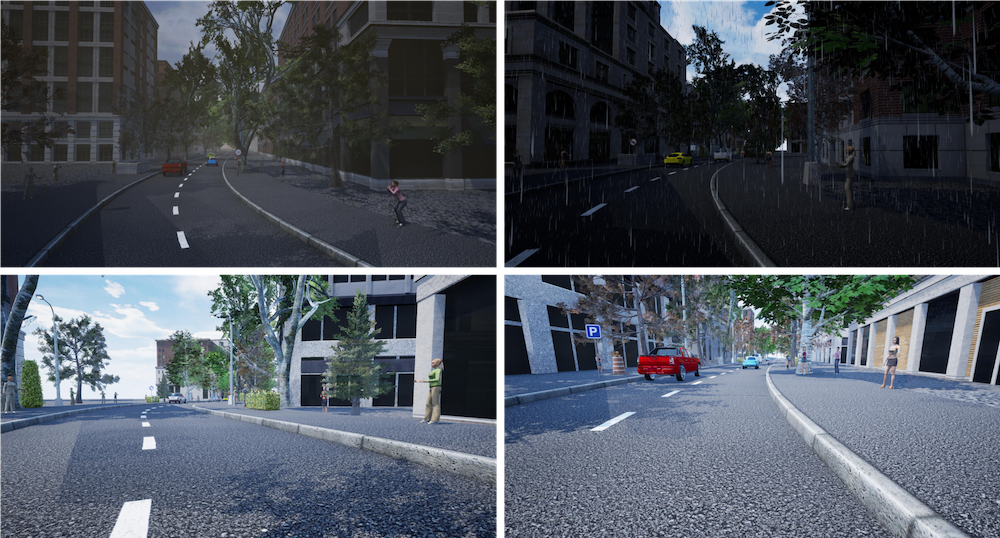}
    \caption{Samples of the training database.}
    \label{fig:speed}
\end{figure}

\textbf{Testing data:} RGB images with a total size of around 16 GB.

\textbf{Expected output on testing data:} traffic sign locations (x1,y1,x2,y2,x3,y3,x4,y4) and the classes of the detected traffic signs (0~20). 

\textbf{Evaluation metric: } Intersection over Union (IOU) between ground truth and predicted bounding box will be used to evaluate detection accuracy. Detections' IOU should be larger than 0.9 to be eligible for recognition evaluation.  

\begin{equation}
IOU=\frac{\text { area of overlap }}{\text { area of union }}    
\end{equation}

The F1 score will be used as the evaluation metric for traffic sign recognition. The F1 score is defined as
\begin{equation} \label{eq:1}
F_1=\frac{2 \text {precision} * \text {recall}}{\text {precision}+\text {recall}}
\end{equation}

\section{Method: RetinaNet}
\begin{figure}[H]
    \centering
    \includegraphics[width=1\textwidth]{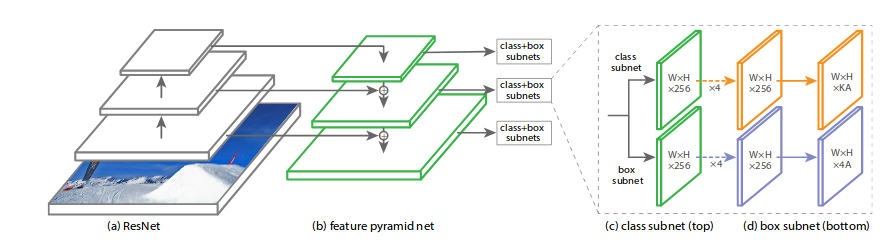}
    \caption{RetinaNet network architecture \citep{lin2017focal}.}
    \label{fig:retina}
\end{figure}

A method based on RetinaNet \citep{lin2017focal} was developed to detect and recognize traffic signs in this study. RetinaNet is a single, unified network composed of a backbone network and two task-specific subnetworks, as shown in Figure \ref{fig:retina}. The backbone is responsible for computing a convolution feature map over an entire input image and is an off-the-self convolution network. The first subnet performs classification on the backbones output; the second subnet performs convolution bounding box regression \citep{WinNT}. We will describe each important component of this method below.

\textbf{Image Preprocessing: Cropping}\\
The original sizes of the images in this study are 3200 $\times$ 1800, with traffic sign being about $50 \times 50$ large. Directly using the large images as training inputs has two downsides: 1) the training speed is slow because we had to use small batch sizes (like 4) to avoid out-of-memory issues; 2) relative to the whole big image, the traffic sign are rather small and hard to detect (we have failed to detect anything due to this). 

\begin{figure}[H]
    \centering
    \includegraphics[width=1\textwidth]{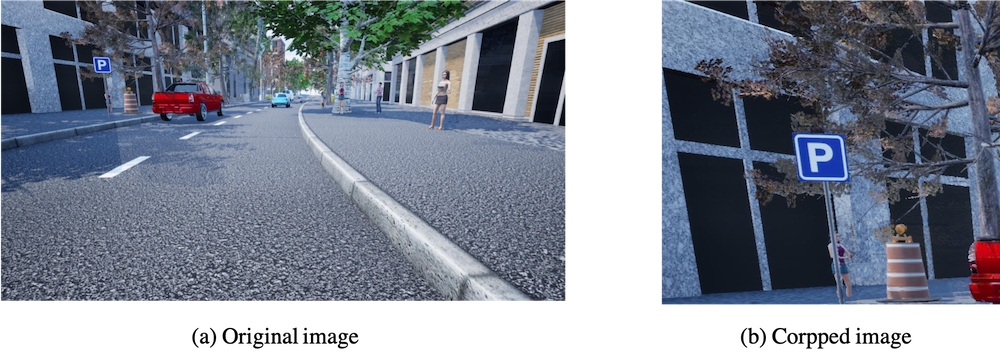}
    \caption{Illustration of image cropping.}
    \label{fig:crop}
\end{figure}

To overcome this issue, we used cropping technique to let the training network focus on samll regions of the whole image. Specifically, for training and validation data, we take the target center (derived from data labels) as the origin, then extend 200 pixels to the left, right, top, and bottom, resulting in an area with $400 \times 400$ size. The extended area is cropped and selected as training and validation images, as shown in Figure \ref{fig:crop}. For testing data, we first randomly crop the images into $400 \times 400$ sub images. Then detect and recognize traffic signs based on these randomly cropped sub images. Finally, we combine the results from multiple sub images and generate a overall prediction result.

It should be noted that when doing image cropping for train and validation images, we also need to update the bounding box coordinates accordingly to make sure they use the cropped images as references. Besides image cropping, regular image augmentation skills like scaling and flip were also applied to the sub images to avoid over-fitting.

\begin{figure}[H]
    \centering
    \includegraphics[width=0.4\textwidth]{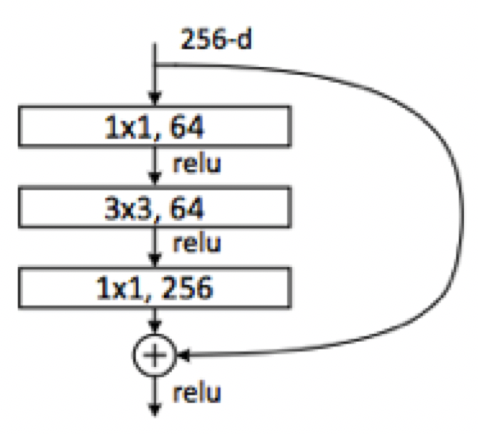}
    \caption{ResNet 101 residual block.}
    \label{fig:res50}
\end{figure}

\textbf{Feature Pyramid Network Backbone}\\
Feature Pyramid Network (FPN) augments a standard convolutional network with a top-down pathway and lateral connections so the network efficiently constructs a rich, multi-scale feature pyramid from a single resolution input image. Each level of the pyramid can be used for detecting objects at a different scale \citep{lin2017focal}. In this study, we used ResNet 101 \citep{he2016deep} as a backbone network for feature extraction. The residual block of the ResNet 101 is shown in Figure \ref{fig:res50}.

\textbf{Anchors}\\
As the original RetinaNet paper, we used anchors with areas of $32^2$ to $512^2$ on pyramid levels P3 to P7, respectively. At each pyramid level we use anchors at three aspect ratios {1:2, 1:1, 2:1}. At each level we add anchors of sizes $\{0.1, 0.2, 0.5, 0.8, 1.0, 1.5\}$ of the original set of 3 aspect ratio anchors, rather than anchors of sizes $\{2^0, 2^{\frac{1}{3}}, 2^{\frac{2}{3}}\}$ used in the original RetinaNet paper, with the purpose of detect more potential traffic signs with different sizes and shapes. 

\textbf{Classification and Box Regression Subnet}\\
For classification and box regression subnet, we used the same setting as the original RetinaNet paper \citep{lin2017focal}. Basically, the classification subnet predicts the probability of object presence at each spatial position for each of the $A$ anchors and $K$ object classes. For each of the $A$ anchors per spatial location, these 4 outputs predict the relative offset between the anchor and the groundtruth box.

\textbf{Network Training}\\
We created a n1-standard-8 (8 vCPUs, 30 GB memory) Virtual Machine on Google Cloud Platform equipped with $2 \times$ NVIDIA Tesla P100 GPUs. Over the total of around 20,000 images, 70\% were used as training data, 15\% were used as validation data, and the remaining 15\% were used as testing data. The hyper parameters were set as follows: learning rate = 0.0001, batch size = 36 (multiple GPUs), number of steps per epoch = 2000, and number of epochs to train = 30. The final weight of our best model trained on the traffic sign dataset can be accessed at https://drive.google.com/file/d/1lzpNg9XVYLFMKZBE5gizEDYkJS9wkmT6/view?usp=sharing.

\section{Results}
The final F1 score of our model on the testing data are 0.923. The average processing time for every picture is around 0.15 seconds. Some true and false detection and recognition results are shown below.

\begin{figure}[H]
    \centering
    \begin{minipage}[t]{0.48\textwidth}
    \centering
    \includegraphics[width=6cm,height=6cm]{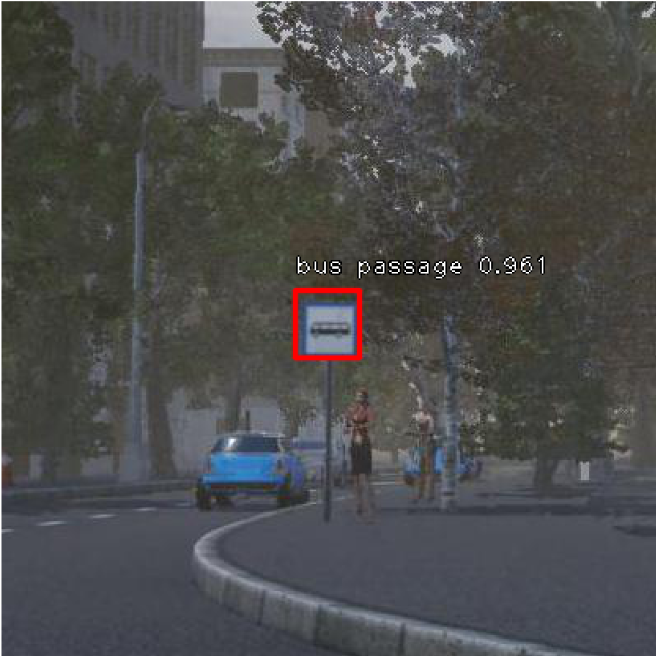}
    \caption{Bus Passage Detection.}
    \end{minipage}
    \begin{minipage}[t]{0.48\textwidth}
    \centering
    \includegraphics[width=6cm,height=6cm]{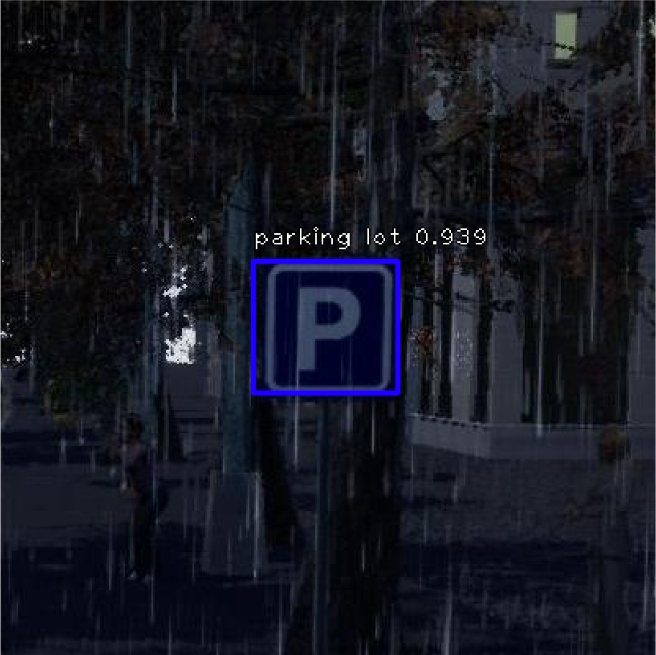}
    \caption{Parking Lot Detection.}
    \end{minipage}
\end{figure}

\begin{figure}[H]
    \centering
    \begin{minipage}[t]{0.48\textwidth}
    \centering
    \includegraphics[width=6cm,height=6cm]{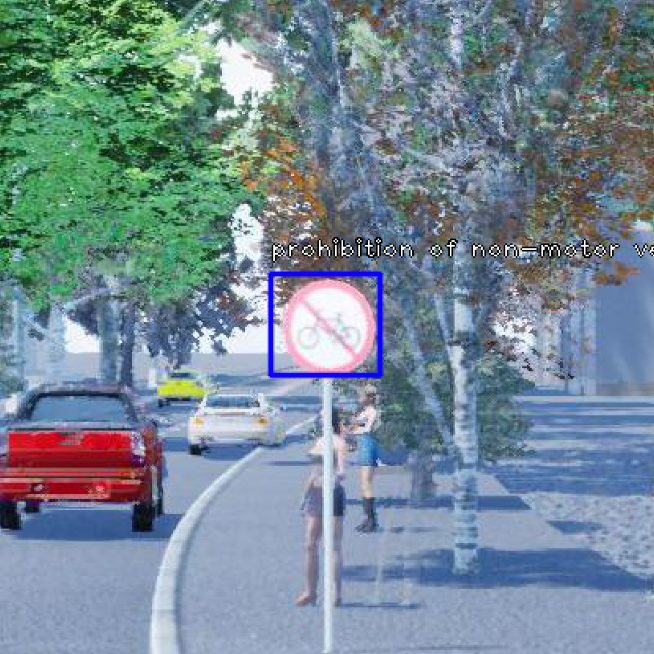}
    \caption{Prohibition of Non-Motor Detection.}
    \end{minipage}
    \begin{minipage}[t]{0.48\textwidth}
    \centering
    \includegraphics[width=6cm,height=6cm]{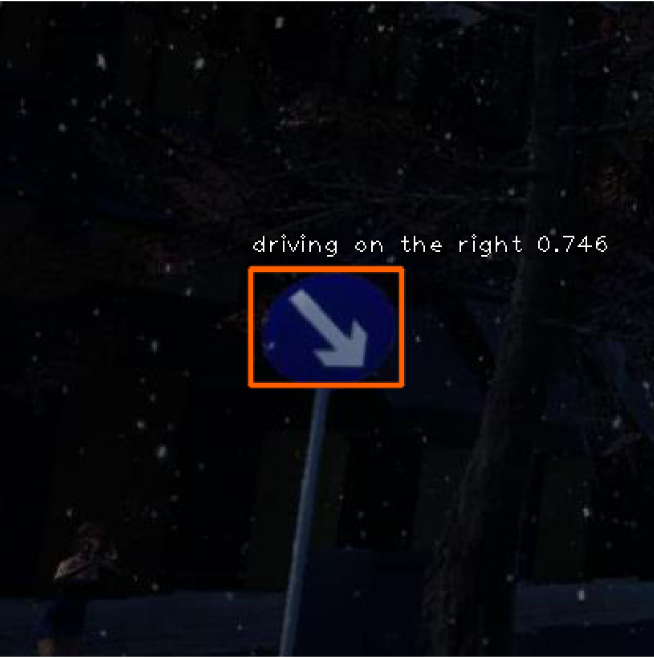}
    \caption{Driving on the Right Detection.}
    \end{minipage}
\end{figure}

\begin{figure}[H]
    \centering
    \begin{minipage}[t]{0.48\textwidth}
    \centering
    \includegraphics[width=6cm,height=6cm]{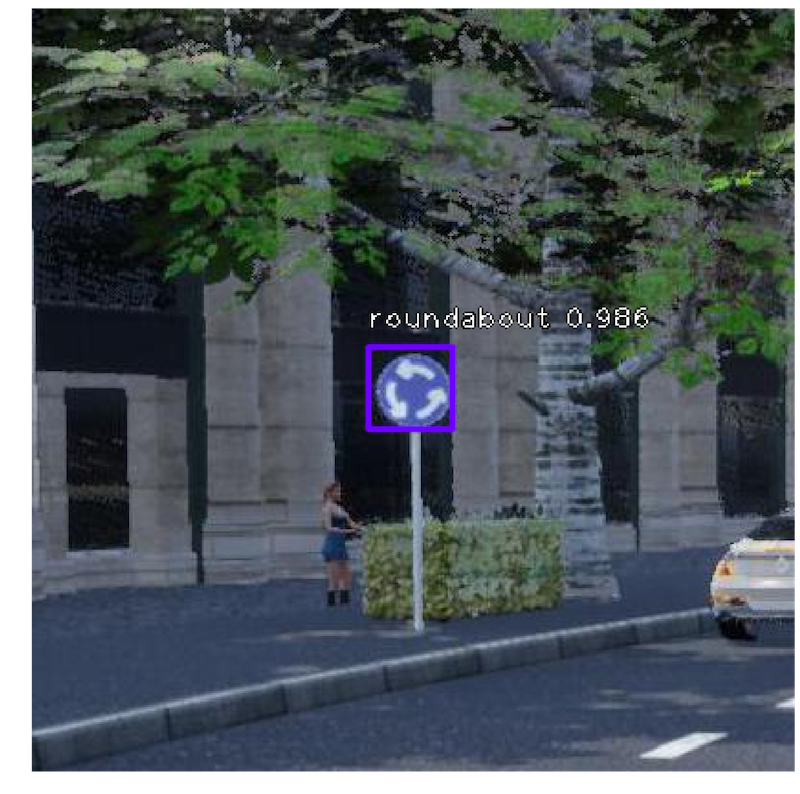}
    \caption{Roundabout detection.}
    \end{minipage}
    \begin{minipage}[t]{0.48\textwidth}
    \centering
    \includegraphics[width=6cm,height=6cm]{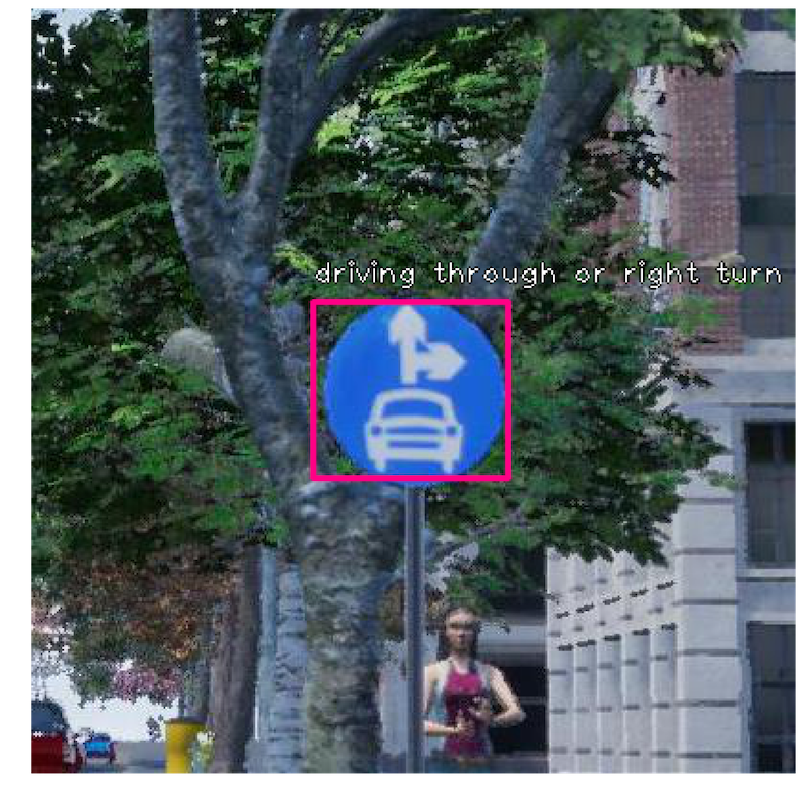}
    \caption{Driving through or right turn detection.}
    \end{minipage}
\end{figure}

\begin{figure}[H]
    \centering
    \begin{minipage}[t]{0.48\textwidth}
    \centering
    \includegraphics[width=6cm,height=6cm]{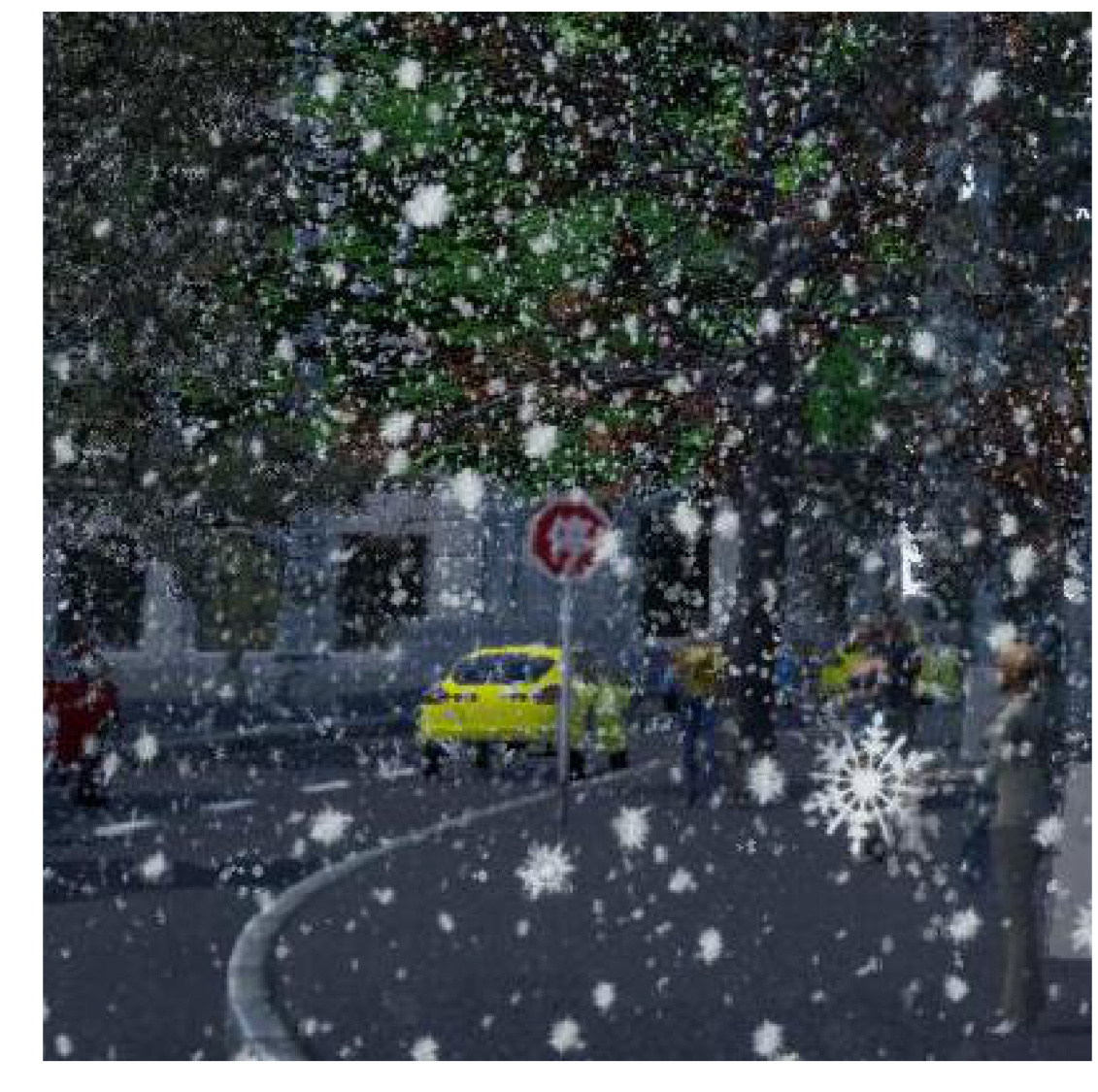}
    \caption{Failed to detect example.}
    \end{minipage}
    \begin{minipage}[t]{0.48\textwidth}
    \centering
    \includegraphics[width=6cm,height=6cm]{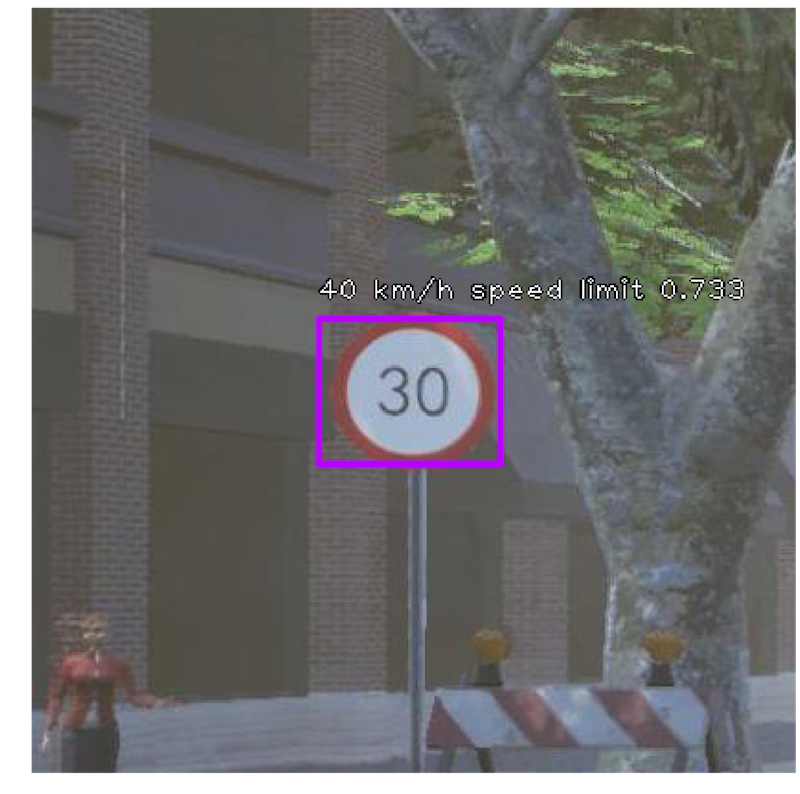}
    \caption{Wrong recognition example.}
    \end{minipage}
\end{figure}

\section{Discussion}
The study developed a traffic sign detection and recognition algorithm based on RetinaNet. Specifically, two main aspects were revised to improve the detection of traffic signs: image cropping to address the issue of large image and small traffic signs; and using more anchors with various scales to detect traffic signs with different sizes and shapes. The algorithm' strength is it performs extremely well under good illumination and weather condition, with reasonable speed. And the weakness is that it sometimes fails to detect object under bad weather conditions like snow. And it fails to distinguish between speed limits signs with different limit values. 
Some general discussion:\\
\begin{itemize}
    \item Image cropping is important for this problem. Without it, we can not detect anything.
    \item We have also tried Yolo v3. It can detect objects well but fail to recognize well.
    \item Setting an proper anchor size and shape is vital for object detection. 
\end{itemize}

\section{Future Work}
The work can be further improved in the following aspects:\\

\begin{itemize}
    \item Do random cropping for training and validation data to improve the model's performance.
    \item Deal with class imbalance so that the minority class can also be well detected and recognized.
    \item Do shape detection (like circle detection) first to localize the traffic signs.
    \item Play with the anchor size and shapes so that we do not need to do image cropping.
    \item Try methods in the field of small objects detection, like .
    \item Try more detection algorithms.
    \item Try detection by segmentation, like Mask-RCNN \citep{he2017mask}.
    \item Try recently emerging anchor-free detection methods so that we can better deal with scales, like \cite{chen2019dubox}.
\end{itemize}

\section{Acknowledgement}
Thanks for Linda Shapiro's great course and the suggestion that we should focus on local region with traffic signs rather than a whole image. Thanks for Bendita and Deepali's suggestions and feedback.

\medskip

\small

\bibliography{name}

\begin{thebibliography}{12}
\providecommand{\natexlab}[1]{#1}
\providecommand{\EM}{\em}
\providecommand{\RNtxt}{\relax}
\RNtxt{}

\bibitem[Arcos-Garcia et~al.(2018)A.~Arcos-Garcia, J.~A.
  {\'A}lvarez-Garc{\'\i}a, L.~M. Soria-Morillo]{arcos2018evaluation}
{\EM Arcos-Garcia Alvaro, {\'A}lvarez-Garc{\'\i}a Juan~A, Soria-Morillo
  Luis~M}.
\newblock Evaluation of Deep Neural Networks for traffic sign detection systems
  \allowbreak\newblock// Neurocomputing. 2018. 316. 332--344.

\bibitem[Chen et~al.(2019)S.~Chen, J.~Li, C.~Yao, W.~Hou, S.~Qin, W.~Jin,
  X.~Tang]{chen2019dubox}
{\EM Chen Shuai, Li~Jinpeng, Yao Chuanqi, Hou Wenbo, Qin Shuo, Jin Wenyao, Tang
  Xu}.
\newblock DuBox: No-Prior Box Objection Detection via Residual Dual Scale
  Detectors \allowbreak\newblock// arXiv preprint arXiv:1904.06883. 2019.

\bibitem[Dai et~al.(2016)J.~Dai, Y.~Li, K.~He, J.~Sun]{dai2016r}
{\EM Dai Jifeng, Li~Yi, He~Kaiming, Sun Jian}.
\newblock R-fcn: Object detection via region-based fully convolutional networks
  \allowbreak\newblock// Advances in neural information processing systems.
  2016.  379--387.

\bibitem[He et~al.(2017)K.~He, G.~Gkioxari, P.~Doll{\'a}r,
  R.~Girshick]{he2017mask}
{\EM He~Kaiming, Gkioxari Georgia, Doll{\'a}r Piotr, Girshick Ross}.
\newblock Mask r-cnn \allowbreak\newblock// Proceedings of the IEEE
  international conference on computer vision. 2017.  2961--2969.

\bibitem[He et~al.(2016)K.~He, X.~Zhang, S.~Ren, J.~Sun]{he2016deep}
{\EM He~Kaiming, Zhang Xiangyu, Ren Shaoqing, Sun Jian}.
\newblock Deep residual learning for image recognition \allowbreak\newblock//
  Proceedings of the IEEE conference on computer vision and pattern
  recognition. 2016.  770--778.

\bibitem[Howard et~al.(2017)A.~G. Howard, M.~Zhu, B.~Chen, D.~Kalenichenko,
  W.~Wang, T.~Weyand, M.~Andreetto, H.~Adam]{howard2017mobilenets}
{\EM Howard Andrew~G, Zhu Menglong, Chen Bo, Kalenichenko Dmitry, Wang Weijun,
  Weyand Tobias, Andreetto Marco, Adam Hartwig}.
\newblock Mobilenets: Efficient convolutional neural networks for mobile vision
  applications \allowbreak\newblock// arXiv preprint arXiv:1704.04861. 2017.

\bibitem[Jay(2018)P.~Jay]{WinNT}
{\EM Jay Prakash}.
\newblock The intuition behind RetinaNet. 2018.

\bibitem[Lin et~al.(2017)T.-Y. Lin, P.~Goyal, R.~Girshick, K.~He,
  P.~Doll{\'a}r]{lin2017focal}
{\EM Lin Tsung-Yi, Goyal Priya, Girshick Ross, He~Kaiming, Doll{\'a}r Piotr}.
\newblock Focal loss for dense object detection \allowbreak\newblock//
  Proceedings of the IEEE international conference on computer vision. 2017.
  2980--2988.

\bibitem[Liu et~al.(2016)W.~Liu, D.~Anguelov, D.~Erhan, C.~Szegedy, S.~Reed,
  C.-Y. Fu, A.~C. Berg]{liu2016ssd}
{\EM Liu Wei, Anguelov Dragomir, Erhan Dumitru, Szegedy Christian, Reed Scott,
  Fu~Cheng-Yang, Berg Alexander~C}.
\newblock Ssd: Single shot multibox detector \allowbreak\newblock// European
  conference on computer vision. 2016.  21--37.

\bibitem[Redmon, Farhadi(2018)J.~Redmon, A.~Farhadi]{redmon2018yolov3}
{\EM Redmon Joseph, Farhadi Ali}.
\newblock Yolov3: An incremental improvement \allowbreak\newblock// arXiv
  preprint arXiv:1804.02767. 2018.

\bibitem[Ren et~al.(2015)S.~Ren, K.~He, R.~Girshick, J.~Sun]{ren2015faster}
{\EM Ren Shaoqing, He~Kaiming, Girshick Ross, Sun Jian}.
\newblock Faster r-cnn: Towards real-time object detection with region proposal
  networks \allowbreak\newblock// Advances in neural information processing
  systems. 2015.  91--99.

\bibitem[Szegedy et~al.(2017)C.~Szegedy, S.~Ioffe, V.~Vanhoucke, A.~A.
  Alemi]{szegedy2017inception}
{\EM Szegedy Christian, Ioffe Sergey, Vanhoucke Vincent, Alemi Alexander~A}.
\newblock Inception-v4, inception-resnet and the impact of residual connections
  on learning \allowbreak\newblock// Thirty-First AAAI Conference on Artificial
  Intelligence. 2017.

\end{thebibliography}

\end{document}